\documentclass[10pt,twocolumn,letterpaper]{article}

\usepackage[pagenumbers]{cvpr} 

\usepackage{graphicx}
\usepackage{amsmath}
\usepackage{amssymb}
\usepackage{booktabs}
\usepackage{multirow}
\usepackage{bbding}
\usepackage[pagebackref,breaklinks,colorlinks]{hyperref}

\usepackage[capitalize]{cleveref}
\crefname{section}{Sec.}{Secs.}
\Crefname{section}{Section}{Sections}
\Crefname{table}{Table}{Tables}
\crefname{table}{Tab.}{Tabs.}
\Crefname{figure}{Figure}{Figures}
\crefname{figure}{Fig.}{Figs.}

\begin{document}

\title{CrowdRec: 3D Crowd Reconstruction from Single Color Images}

\author{Buzhen Huang\hspace{5mm} Jingyi Ju\hspace{5mm} Yangang Wang\\%
\\
Southeast University, China
}

\twocolumn[{%
\renewcommand\twocolumn[2][]{#1}%
\maketitle
\thispagestyle{empty}
\begin{center}
   \centering
   \includegraphics[width=1.0\textwidth]{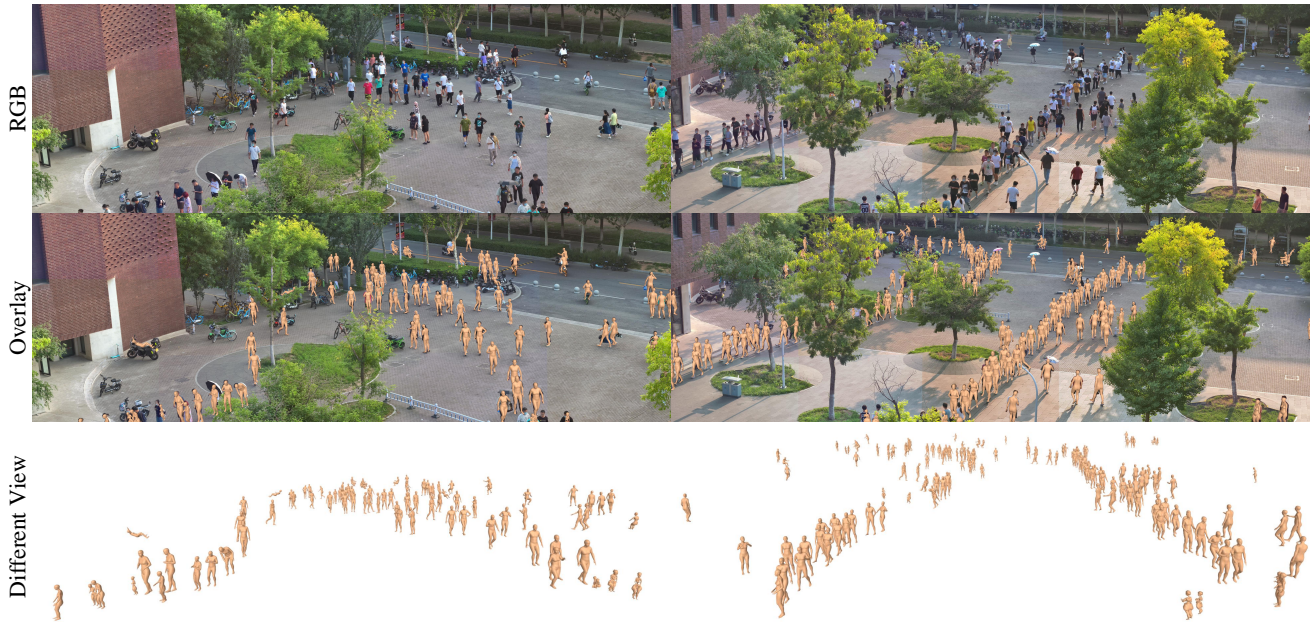}
   \vspace{-5mm}
   \captionof{figure}{Our method reconstructs 3D crowds with accurate body poses, shapes, and absolute positions from a monocular color image.}
   \label{fig:teaser}
\end{center}
}]

\begin{abstract}
   This is a technical report for the GigaCrowd challenge. Reconstructing 3D crowds from monocular images is a challenging problem due to mutual occlusions, server depth ambiguity, and complex spatial distribution. Since no large-scale 3D crowd dataset can be used to train a robust model, the current multi-person mesh recovery methods can hardly achieve satisfactory performance in crowded scenes. In this paper, we exploit the crowd features and propose a crowd-constrained optimization to improve the common single-person method on crowd images. To avoid scale variations, we first detect human bounding-boxes and 2D poses from the original images with off-the-shelf detectors. Then, we train a single-person mesh recovery network using existing in-the-wild image datasets. To promote a more reasonable spatial distribution, we further propose a crowd constraint to refine the single-person network parameters. With the optimization, we can obtain accurate body poses and shapes with reasonable absolute positions from a large-scale crowd image using a single-person backbone. The code will be publicly available at~\url{https://github.com/boycehbz/CrowdRec}.

\end{abstract}


\begin{figure*}
    \begin{center}
    \includegraphics[width=1.0\linewidth]{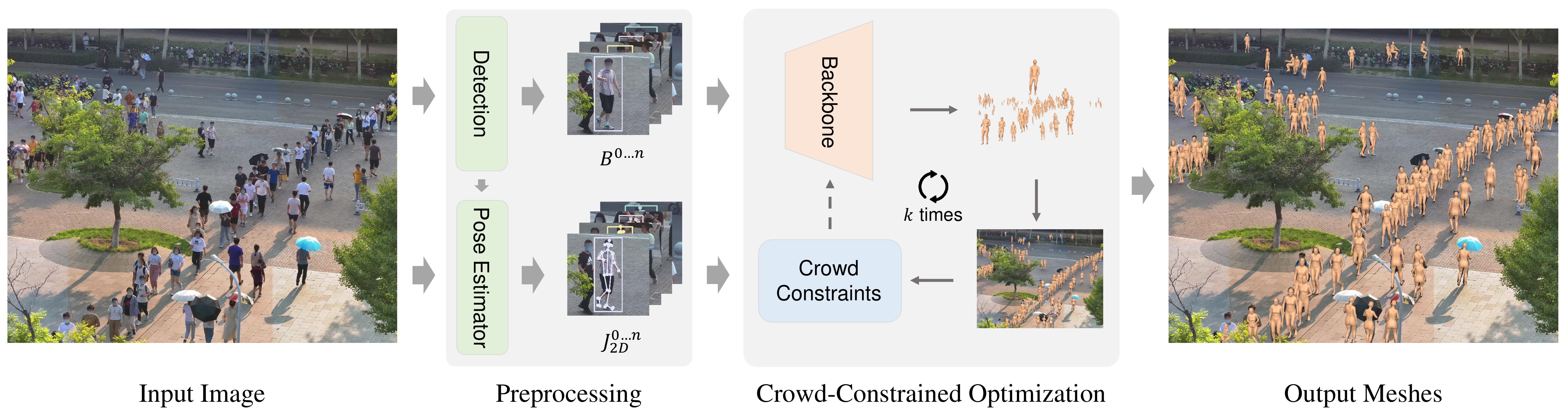}
    \end{center}
    \vspace{-6mm}
    \caption{Overview of our method. We first detect human bounding-boxes and corresponding 2D poses with off-the-shelf detectors. Then, a single-person backbone is used to regress 3D human models from the cropped image patches. We further optimize the backbone network parameters with crowd constraints. After several iterations, accurate 3D crowds can be obtained from the optimized network.}
\label{fig:pipeline}
\vspace{-4mm}
\end{figure*}

\section{Method}\label{sec:Method}
Given a large-scale crowd image~(\eg, gigapixel-level image~\cite{wang2020panda}), our method reconstructs 3D human models with accurate body pose, shape, and absolute position. We first detect human bounding-boxes and 2D poses with off-the-shelf detectors~(\cref{sec:Preliminaries}). Then, a single-person mesh recovery network is trained as a backbone on in-the-wild datasets~(\cref{sec:Backbone}). Due to the depth ambiguity and shape-depth coupling, successively regressing each human from the crowd image can not produce reasonable spatial distribution. Therefore, we further propose a crowd-constrained optimization to exploit crowd features to refine the network~(\cref{sec:Optimization}). After several iterations, the accurate crowd meshes can be obtained from the finetuned network.

\subsection{Preliminaries}\label{sec:Preliminaries}
\noindent\textbf{Human detection.} Since bottom-up methods~\cite{sun2021monocular,sun2022putting} confront server scale variations, we propose a top-down strategy to reconstruct the crowd from a large image. We first adopt YOLOX~\cite{ge2021yolox} to detect each human from the image. To achieve better performance on large-scale crowded scenes, we use pre-trained weights on MOT17~\cite{milan2016mot16}, CrowdHuman~\cite{shao2018crowdhuman}, Cityperson~\cite{zhang2017citypersons} and ETHZ~\cite{ess2008mobile} datasets from ByteTrack~\cite{zhang2022bytetrack}. Thus, the model is more robust to occlusions and scale variations. We set the detection threshold to 0.23 in our method.

\noindent\textbf{2D pose estimation.} We further use 2D poses as an additional constraint to promote more accurate body poses. To keep the original high-resolution, we use the bounding-box from YOLOX to crop image patches and estimate 2D pose for each instance with AlphaPose~\cite{fang2022alphapose}. Other 2D pose detectors~\cite{sun2019deep,cao2017realtime} are also feasible in our framework. The predicted poses are then mapped to the original image coordinates.

\subsection{Single-Person Backbone Pre-Training}\label{sec:Backbone}
Since no large-scale 3D crowd dataset can be used for training a multi-person mesh recovery model, our strategy is to use a single-person backbone with crowd constraints to achieve crowd reconstruction. Therefore, we can fully utilize single-person datasets to train a robust model for in-the-wild images. We crop the image patches with the human bounding-boxes and then extract human features with the HRNet network~\cite{sun2019deep}. To incorporate the location information into the regression, we follow CLIFF~\cite{li2022cliff} to concatenate the bounding-box information with the extracted features. Then we use 3 heads to regress SMPL~\cite{loper2015smpl} pose $\theta$, shape $\beta$, and camera $[f_c, t_x, t_y]$ parameters. The predicted camera can be further transformed into absolute translation:
\begin{equation}
    t_X =t_x + \frac{2 c_x}{d f_c}, t_Y =t_y + \frac{2 c_y}{d f_c}, t_Z =\frac{2 f}{d f_c}, \label{equ:translation}
\end{equation}
where $t = [t_X, t_Y, t_Z]$ is the translation. $(c_x, c_y)$ is the bounding-box location relative to the original image center, and $d$ is its size. $f$ is the length of the diagonal line of the original image, which is used to approximate the focal length for uncalibrated images.

We use the following constraints to train the model.
\begin{equation}
    \mathcal{L} = \lambda_{1} \mathcal{L}_{\text{reproj}} + \lambda_{2} \mathcal{L}_{\text{smpl}} + \lambda_{3} \mathcal{L}_{\text{joint}} + \lambda_{4} \mathcal{L}_{\text{verts}},\label{equ:loss}
\end{equation}
where $\lambda_{1}=5.0$, $\lambda_{2}=5.0$, $\lambda_{3}=1.0$, and $\lambda_{4}=0.1$ are loss weights. With the predicted global translation, we project the joints to the original image plane and calculate the reprojection error. Specifically, we add the translation $t$ to the SMPL 3D joint positions $J_{3D}$ and calculate the loss with following function:
\begin{equation}
    \mathcal{L}_{\text{reproj}} = \| \Pi\left(J_{3 D}+ t\right) - \hat{J_{2D}} \|_2^2,\label{equ:reproj}
\end{equation}
where $\Pi$ projects the 3D joints to 2D image with focal length $f$ and image center. $\hat{J_{2D}}$ is ground-truth 2D pose. We normalize the reprojection error with the corresponding bounding-box.

The SMPL parameters and 3D joint positions are also used for supervision:
\begin{equation}
    \mathcal{L}_{\text{smpl}} = \| [\beta, \theta] - [\hat{\beta}, \hat{\theta}] \|_2^2.
\end{equation}
\begin{equation}
    \mathcal{L}_{\text{joint}} = \| J_{3D} - \hat{J_{3D}} \|_2^2.
\end{equation}
The $\hat{\beta}$, $\hat{\theta}$, and $\hat{J_{3D}}$ are ground-truth annotations. We also use vertex error as supervision:
\begin{equation}
    \mathcal{L}_{\text{verts}} = \| V_{3D} - \hat{V_{3D}} \|_2^2,
\end{equation}
where $V_{3D}$ is the vertex position generated with SMPL parameters according to its kinematic tree.

\begin{figure*}
    \begin{center}
    \includegraphics[width=1.0\linewidth]{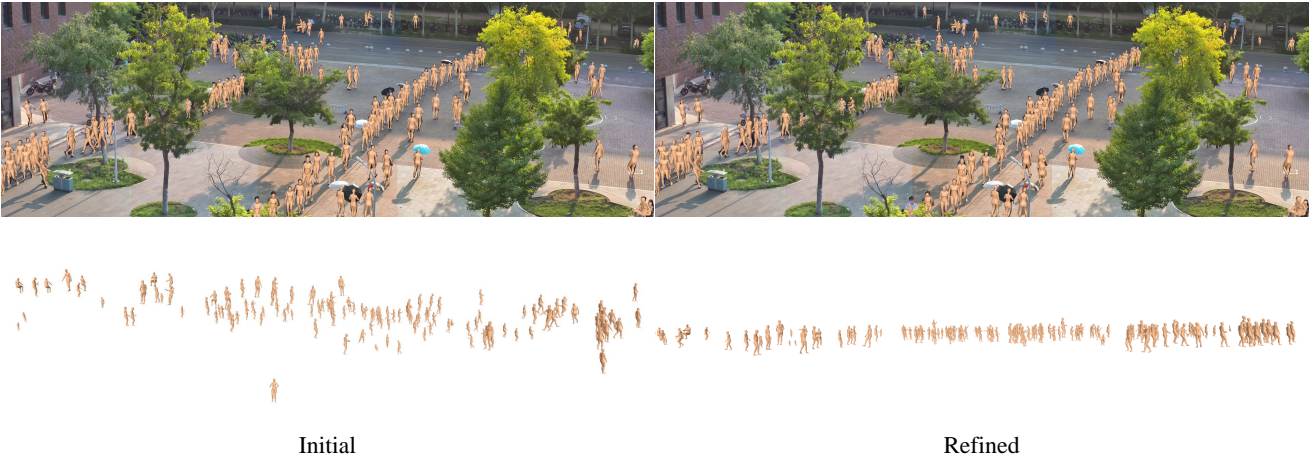}
    \end{center}
    \vspace{-7mm}
    \caption{The output meshes from the single-person backbone have an unreasonable spatial distribution. With the crowd-constrained optimization, the refined meshes have more accurate poses, shapes, and absolute positions.}
\label{fig:comparison3}
\vspace{-6mm}
\end{figure*}

\subsection{Crowd-Constrained Optimization}\label{sec:Optimization}
Human mesh recovery from monocular images is an ill-posed problem, and the skeletal pose, body shape, and global depth couple with each other. Successively regressing each person with the trained single-person model always leads to unreasonable spatial distribution. Thus, we exploit crowd features to optimize the results. It is common that the root of the people in crowded scenes are on limited adjacent planes. For example, the people should be on the same plane when the scene is a playground. However, estimating the ground plane requires complex calibration procedures and specific patterns. Following \cite{grouprec}, we directly predict ground plane normal with human features. When all people in the scene are regressed with the single-person backbone, we approximate the normal direction with human head and ankle keypoints.
\begin{equation}
    l = \frac{1}{N} \sum_{n=1}^N \frac{J_{\text{top}}^{n} - J_{\text{bottom}}^{n}}{\|J_{\text{top}}^{n} - J_{\text{bottom}}^{n} \|}.
\end{equation}
$J_{\text{top}}$ is 3D keypoint on the head, and $J_{\text{bottom}}$ is the middle point of two ankle keypoints. $N$ is the total number of people in the scene. The assumption is valid in most common crowded scenes. We then project the root position of all people to the normal direction and penalize the standard deviation.
\begin{equation}
    \mathcal{L}_{\text{crowd}} = std(J_{root} \cdot l),\label{equ:crowd}
\end{equation}
where $std(\cdot)$ denotes standard deviation, and $J_{root} \in \mathbb{R}^{N \times 3}$ is the root positions of all people in the image. $(\cdot)$ means dot product. We require the estimated crowd to have a minor crowd error.

Therefore, we fed a batch of image patches into the backbone, and further apply the additional constraints on the regressed meshes. 
\begin{equation}
    \mathop{\arg\min}_{\theta_\pi} \mathcal{L}= \lambda_{5} \mathcal{L}_{\text{crowd}} + \lambda_{6} \mathcal{L}_{\text{keyp}}+\mathcal{L}_{\text{init}},
\end{equation}
It is noted that we optimize network parameters of the single-person backbone $\theta_\pi$ rather than the output SMPL parameters since the network parameters contain prior knowledge of human pose. The optimized results are also forced to fit predicted 2D poses. 
\begin{equation}
    \mathcal{L}_{\text{keyp}} = \frac{1}{N} \sum_{n=1}^N \| \Pi\left(J_{3 D}^n+ t^n\right) - {J_{2D}^n}^\prime \|_2^2,\label{equ:keyp}
\end{equation}
where ${J_{2D}}^\prime$ is predicted 2D pose. The final results should also be similar to initial prediction.
\begin{equation}
    \mathcal{L}_{\text{init}} = \frac{1}{N} \sum_{n=1}^N \left( \lambda_{7} \mathcal{L}_{\text{crowd}} \|\beta^n - \beta^n_{init} \|_2^2 + \lambda_{8} \|\theta^n - \theta^n_{init} \|_2^2 \right).
\end{equation}
$\theta_{init}$ and $\beta_{init}$ are initial pose and shape parameters predicted from the single-person backbone. In our method, we adopt the hyperparameters of $\lambda_{5}=0.001$, $\lambda_{6}=5.0$, $\lambda_{7}=0.001$, and $\lambda_{8}=5.0$.

When the total loss is convergent, we output the SMPL parameters from the optimized network.

\section{Experiments}\label{sec:Experiments}

\begin{table*}
    \begin{center}
        \resizebox{0.55\linewidth}{!}{
            \begin{tabular}{r|c c c c c}
            \noalign{\hrule height 1.5pt}
            \begin{tabular}[l]{l}\multirow{1}{*}{Method}\end{tabular}
                &PA-PPDS$\uparrow$ &SS$\uparrow$ &PCOD$\uparrow$ &OKS$\uparrow$ &RP$\downarrow$  \\
            \noalign{\hrule height 1pt}
            \hline \hline
            80 \textit{iter} + 0.45 \textit{thre} &0.8163 &0.8982 &0.8788 &0.7466 &0.000 \\
            200 \textit{iter} + 0.45 \textit{thre} &0.8245 &0.9559 &0.8786 &0.7642 &0.000 \\
            260 \textit{iter} + 0.24 \textit{thre} &0.8378 &0.9664 &0.8979 &0.7792 &0.001 \\
            260 \textit{iter} + 0.23 \textit{thre} &0.8385 &0.9665 &0.8990 &0.7801 &0.001 \\
            \noalign{\hrule height 1.5pt}
            \end{tabular}
        }
\vspace{-2mm}
\caption{We analyze the effect of different optimization iterations and detection thresholds. Our method achieves better performance with more iterations. \textit{iter} means iterations, and \textit{thre} denotes threshold.}
\label{tab:crowd}
\end{center}
\vspace{-5mm}
\end{table*}

\begin{figure*}
    \begin{center}
    \includegraphics[width=1.0\linewidth]{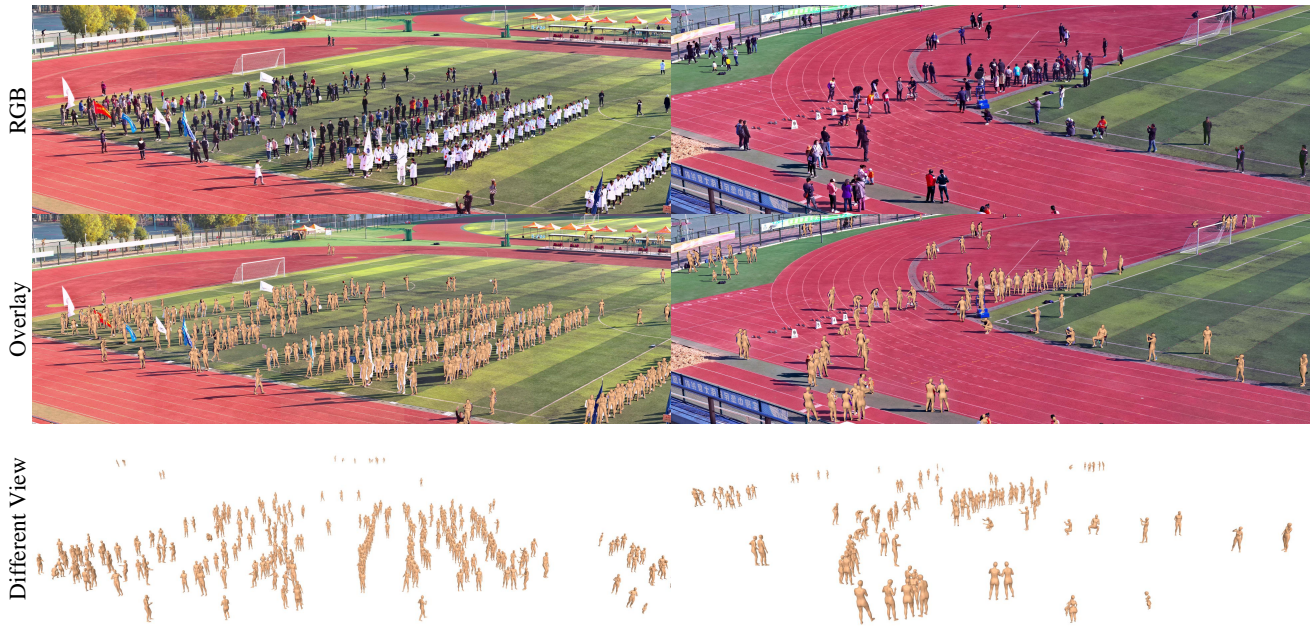}
    \end{center}
    \vspace{-6mm}
    \caption{Our method also performs well on crowded scenes with server occlusions.}
\label{fig:comparison_1}
\vspace{-6mm}
\end{figure*}

\subsection{Dataset}
\noindent\textbf{Training data.} We follow common single-person methods to use Human3.6M~\cite{ionescu2013human3}, COCO~\cite{lin2014microsoft}, MPII~\cite{andriluka20142d} to train the backbone. For Human3.6M, we use S1, S5, S6, S7, and S8 with SMPL annotations to train the model. We also use pseudo ground-truth from CLIIF~\cite{li2022cliff} for COCO and MPII. With the in-the-wild image datasets, the single-person backbone can be robust to different scenes and occlusions.

\noindent\textbf{Testing data.} We use the standard test set of GigaCrowd~\cite{DSGnet} for evaluation. It contains 298 frames in 4 large-scale scenes with crowds. We evaluate the pose accuracy and spatial positions in this challenging benchmark.

\subsection{Metric}
We follow the GigaCrowd challenge to use PA-PPDS, SS, OKS, PCOD, and RP to evaluate our method. The detailed formulations of these metrics can be found at their website.\footnote{https://www.gigavision.cn/track/track?nav=GigaCrowd\&type=nav}

\subsection{Implementation Details}
We implement the network using PyTorch. The single-person backbone is trained on an NVIDIA GeForce RTX 2080 Ti GPU for 105 epochs with a batch size of 32 and a learning rate of 1e-4. For the crowd-constrained optimization, we use the pre-trained single-person backbone as initial weights and optimize the network parameters with the constraints. Since the number of people in a single gigapixel image is large, we split the people into several batches. Each batch has 50 people. We iteratively optimize the network parameters from the initial weights for 260 times with AdamW optimizer and cosine annealing decays~\cite{loshchilov2017decoupled}. The initial learning rate for the optimization is 1e-5. The optimization is run with an NVIDIA GeForce RTX 3090 GPU.

\subsection{Results}
We show the results on GigaCrowd dataset in \cref{tab:crowd}. We first use the threshold of 0.45 for the detection and refine the network parameters with 80 iterations. YOLOX with a 0.45 threshold can detect most of the people in the scene. However, it still misses some instances in the distance. Thus, we turn down the threshold to 0.23. Although the people may be redundant, it achieves the best performance. In addition, the iterations strongly affect the performance. Since simultaneously optimizing a large number of people is a highly non-convex problem, the optimization requires more iterations for annealing to adjust the learning rate to find the global minima.

We further compare our method with and without the crowd-constrained optimization in \cref{fig:comparison3}. Due to the shape-depth coupling, the regressed human meshes from the initial single-person backbone have an unreasonable spatial distribution. Although the overlay image is correct, we found that people do not stand in the same plane from the side view. In contrast, the refined meshes with the crowd-constrained optimization are more reasonable. More results in \cref{fig:teaser} and \cref{fig:comparison_1} further demonstrate the great performance of our method in crowded scenes.

\section{Conclusion}\label{sec:conclusion}
In this paper, we propose a crowd-constrained optimization to enable the single-person method to reconstruct accurate body poses and shapes with reasonable absolute positions on large-scale crowded scenes. The constraints are constructed with only human semantics and are valid in crowded scenes. Future works can adopt lightweight single-person backbones and other technical strategies to accelerate the optimization.

{\small
\bibliographystyle{ieee_fullname}
\bibliography{egbib}

\begin{thebibliography}{10}\itemsep=-1pt

\bibitem{andriluka20142d}
Mykhaylo Andriluka, Leonid Pishchulin, Peter Gehler, and Bernt Schiele.
\newblock 2d human pose estimation: New benchmark and state of the art
  analysis.
\newblock In {\em Proceedings of the IEEE Conference on computer Vision and
  Pattern Recognition}, pages 3686--3693, 2014.

\bibitem{cao2017realtime}
Zhe Cao, Tomas Simon, Shih-En Wei, and Yaser Sheikh.
\newblock Realtime multi-person 2d pose estimation using part affinity fields.
\newblock In {\em Proceedings of the IEEE conference on computer vision and
  pattern recognition}, pages 7291--7299, 2017.

\bibitem{ess2008mobile}
Andreas Ess, Bastian Leibe, Konrad Schindler, and Luc Van~Gool.
\newblock A mobile vision system for robust multi-person tracking.
\newblock In {\em 2008 IEEE Conference on Computer Vision and Pattern
  Recognition}, pages 1--8. IEEE, 2008.

\bibitem{fang2022alphapose}
Hao-Shu Fang, Jiefeng Li, Hongyang Tang, Chao Xu, Haoyi Zhu, Yuliang Xiu,
  Yong-Lu Li, and Cewu Lu.
\newblock Alphapose: Whole-body regional multi-person pose estimation and
  tracking in real-time.
\newblock {\em IEEE Transactions on Pattern Analysis and Machine Intelligence},
  2022.

\bibitem{ge2021yolox}
Zheng Ge, Songtao Liu, Feng Wang, Zeming Li, and Jian Sun.
\newblock Yolox: Exceeding yolo series in 2021.
\newblock {\em arXiv preprint arXiv:2107.08430}, 2021.

\bibitem{grouprec}
Buzhen Huang, Jingyi Ju, Zhihao Li, and Yangang Wang.
\newblock Reconstructing groups of people with hypergraph relational reasoning.
\newblock In {\em Proceedings of the IEEE/CVF International Conference on
  Computer Vision}, pages 14873--14883, 2023.

\bibitem{ionescu2013human3}
Catalin Ionescu, Dragos Papava, Vlad Olaru, and Cristian Sminchisescu.
\newblock Human3. 6m: Large scale datasets and predictive methods for 3d human
  sensing in natural environments.
\newblock {\em IEEE transactions on pattern analysis and machine intelligence},
  36(7):1325--1339, 2013.

\bibitem{DSGnet}
Kun Li, Wanpeng Li, Sun Xiaokun, and Fang Lu.
\newblock Deep social grouping network for large scenes with multiple subjects.
\newblock 51(8):1287--1301, 2021.

\bibitem{li2022cliff}
Zhihao Li, Jianzhuang Liu, Zhensong Zhang, Songcen Xu, and Youliang Yan.
\newblock Cliff: Carrying location information in full frames into human pose
  and shape estimation.
\newblock In {\em European Conference on Computer Vision}, pages 590--606.
  Springer, 2022.

\bibitem{lin2014microsoft}
Tsung-Yi Lin, Michael Maire, Serge Belongie, James Hays, Pietro Perona, Deva
  Ramanan, Piotr Doll{\'a}r, and C~Lawrence Zitnick.
\newblock Microsoft coco: Common objects in context.
\newblock In {\em European conference on computer vision}, pages 740--755.
  Springer, 2014.

\bibitem{loper2015smpl}
Matthew Loper, Naureen Mahmood, Javier Romero, Gerard Pons-Moll, and Michael~J
  Black.
\newblock Smpl: A skinned multi-person linear model.
\newblock {\em ACM transactions on graphics (TOG)}, 34(6):1--16, 2015.

\bibitem{loshchilov2017decoupled}
Ilya Loshchilov and Frank Hutter.
\newblock Decoupled weight decay regularization.
\newblock {\em arXiv preprint arXiv:1711.05101}, 2017.

\bibitem{milan2016mot16}
Anton Milan, Laura Leal-Taix{\'e}, Ian Reid, Stefan Roth, and Konrad Schindler.
\newblock Mot16: A benchmark for multi-object tracking.
\newblock {\em arXiv preprint arXiv:1603.00831}, 2016.

\bibitem{shao2018crowdhuman}
Shuai Shao, Zijian Zhao, Boxun Li, Tete Xiao, Gang Yu, Xiangyu Zhang, and Jian
  Sun.
\newblock Crowdhuman: A benchmark for detecting human in a crowd.
\newblock {\em arXiv preprint arXiv:1805.00123}, 2018.

\bibitem{sun2019deep}
Ke Sun, Bin Xiao, Dong Liu, and Jingdong Wang.
\newblock Deep high-resolution representation learning for human pose
  estimation.
\newblock In {\em Proceedings of the IEEE/CVF conference on computer vision and
  pattern recognition}, pages 5693--5703, 2019.

\bibitem{sun2021monocular}
Yu Sun, Qian Bao, Wu Liu, Yili Fu, Michael~J Black, and Tao Mei.
\newblock Monocular, one-stage, regression of multiple 3d people.
\newblock In {\em Proceedings of the IEEE/CVF International Conference on
  Computer Vision}, pages 11179--11188, 2021.

\bibitem{sun2022putting}
Yu Sun, Wu Liu, Qian Bao, Yili Fu, Tao Mei, and Michael~J Black.
\newblock Putting people in their place: Monocular regression of 3d people in
  depth.
\newblock In {\em Proceedings of the IEEE/CVF Conference on Computer Vision and
  Pattern Recognition}, pages 13243--13252, 2022.

\bibitem{wang2020panda}
Xueyang Wang, Xiya Zhang, Yinheng Zhu, Yuchen Guo, Xiaoyun Yuan, Liuyu Xiang,
  Zerun Wang, Guiguang Ding, David Brady, Qionghai Dai, and Lu Fang.
\newblock Panda:a gigapixel-level human-centric video dataset.
\newblock In {\em CVPR}, pages 3265--3275, 2020.

\bibitem{zhang2017citypersons}
Shanshan Zhang, Rodrigo Benenson, and Bernt Schiele.
\newblock Citypersons: A diverse dataset for pedestrian detection.
\newblock In {\em Proceedings of the IEEE conference on computer vision and
  pattern recognition}, pages 3213--3221, 2017.

\bibitem{zhang2022bytetrack}
Yifu Zhang, Peize Sun, Yi Jiang, Dongdong Yu, Fucheng Weng, Zehuan Yuan, Ping
  Luo, Wenyu Liu, and Xinggang Wang.
\newblock Bytetrack: Multi-object tracking by associating every detection box.
\newblock In {\em European Conference on Computer Vision}, pages 1--21.
  Springer, 2022.

\end{thebibliography}
}

\end{document}